%% file: main.tex
\documentclass[
    nonacm,
    sigconf
]{acmart}
\AtBeginDocument{%
  }

\setcopyright{acmlicensed}
\copyrightyear{2018}
\acmYear{2018}
\acmDOI{XXXXXXX.XXXXXXX}
\acmConference[Conference acronym 'XX]{Make sure to enter the correct
  conference title from your rights confirmation email}{June 03--05,
  2018}{Woodstock, NY}
\acmISBN{978-1-4503-XXXX-X/2018/06}

\usepackage{booktabs}
\usepackage{cleveref}

\begin{document}

\title{Grounded Event Extraction from SEC 8-K Filings with a Fine-Grained Taxonomy}

\author{Rian Dolphin}
    \authornote{rian@massive.com
    \\
    The processed data and taxonomy are available at:
    \\
    \href{https://massive.com/docs/rest/stocks/filings/8-k-disclosures?utm_source=research&utm_campaign=8k_tags}{massive.com/docs/rest/stocks/filings/8-k-disclosures}}
    \orcid{0000-0002-5607-9948}
    \affiliation{%
  \institution{\href{https://massive.com?utm_source=research&utm_campaign=8k_tags}{Massive.com}}
  \city{Dublin}
  \country{Ireland}
  }

\author{Joe Dursun}
\affiliation{%
  \institution{\href{https://massive.com?utm_source=research&utm_campaign=8k_tags}{Massive.com}}
  \city{Atlanta, Georgia}
  \country{USA}}

\author{Jarrett Blankenship}
\affiliation{%
  \institution{\href{https://massive.com?utm_source=research&utm_campaign=8k_tags}{Massive.com}}
  \city{Atlanta, Georgia}
  \country{USA}}

\author{Katie Adams}
\affiliation{%
  \institution{\href{https://massive.com?utm_source=research&utm_campaign=8k_tags}{Massive.com}}
  \city{Atlanta, Georgia}
  \country{USA}}

\author{Quinton Pike}
\affiliation{%
  \institution{\href{https://massive.com?utm_source=research&utm_campaign=8k_tags}{Massive.com}}
  \city{Atlanta, Georgia}
  \country{USA}}

\renewcommand{\shortauthors}{Dolphin et al.}

\begin{abstract}
Form 8-K filings are the primary channel through which U.S. public companies disclose material events, but the SEC item codes attached to them are coarse: a single item spans routine administrative changes and chief executive departures, and many of the most market-moving disclosures fall into a catch-all item. Large language models make fine-grained labelling feasible at corpus scale, but only if the labels can be traced to the source text and shown to be reliable. We present a two-stage system that tags 8-K disclosures against a three-tier taxonomy of 119 event types. The first stage constrains output to valid taxonomy entries and anchors every tag to a verbatim quote via fuzzy n-gram validation; the second re-grades each cited quote against the category definition to produce a quality score. Applying the system to 292,984 filings from 2022 to 2026 yields 601,088 grounded event tags, which we release. Over 5,125 stratified tags, an LLM judge finds precision rises monotonically with the quality score, from 12\% to 96\%, while unsupported tags fall from 8\% to near zero. Ablation shows the score is calibrated only when assigned in a dedicated second pass. An event study on unsigned abnormal returns confirms, without any language model, that the taxonomy separates economically distinct events sharing an item code.
\end{abstract}




\maketitle

\input{sections/intro}
\input{sections/related}
\input{sections/method}
\input{sections/data}
\input{sections/intrinsic}
\input{sections/market}
\input{sections/exhibits}
\input{sections/conclusion}


\bibliographystyle{ACM-Reference-Format}
\bibliography{references}

\end{document}

%% file: sections/intro.tex
\section{Introduction}
\label{sec:intro}

U.S. public companies must report material corporate events on Form 8-K within four business days. The 32 SEC item codes attached to each filing, such as Item 5.02 for officer and director changes or Item 1.01 for material agreements, are the standard machine-readable event labels: they determine legal filing obligations, organize disclosure databases, and serve as event-type controls in a large empirical literature~\cite{lerman2010,he2020}. The codes were designed as legal filing categories rather than as economic event types, and they are coarse along two dimensions. Firstly, a single item conflates economically distinct events: Item 5.02 covers both the departure of a chief executive and the routine retirement of a single outside director. Secondly, a large share of substantive disclosure carries no informative item at all: Item 8.01 (``Other Events'') is a voluntary catch-all, and we find that it is the modal location of many of the most price-relevant event types, including clinical trial results, merger completions, and regulatory decisions.

Large language models make it feasible to assign fine-grained, economically motivated event labels at corpus scale. Doing so for research and applied use raises two requirements that generic prompting does not meet. The labels must be \emph{constrained}, so that outputs map onto a fixed vocabulary that downstream users can rely on, and they must be \emph{auditable}, so that every label can be traced to specific language in the source document and assessed for reliability without rerunning the model.

This paper presents and evaluates an LLM-based extraction system built around these requirements. The system places a three-tier taxonomy of 119 corporate event types in the prompt of a compact instruction-tuned language model and tags each filing in two stages. The first stage extracts event tags under two structural reliability mechanisms enforced at the output layer: schema validation that rejects tags outside the taxonomy and triggers self-correcting retries, and fuzzy n-gram quote validation that requires every tag to cite a supporting span verifiably present in the filing. The second stage assigns each tag a quality score from 1 to 5 by re-reading its cited quote against the category definition, an explicit grounding check that yields a calibrated reliability dial. We apply the system to 292{,}984 Form 8-K filings from January 2022 through June 2026, producing 601{,}088 validated event tags.

We evaluate the resulting labels along two complementary lines. The intrinsic evaluation uses a stronger language model as a judge over 5{,}125 tags stratified across all 119 event types and all five quality scores, with verdicts that separate a correct tag from one applied to the wrong category and from one whose claimed event is absent from the filing. The quality score is strongly predictive of precision: it rises monotonically from 12\% at the lowest score to 96\% at the top score, while the rate of tags whose claimed event is absent from the filing falls from 8\% to near zero. Keeping only the top score retains 34\% of tags at 96\% precision, and relaxing to score 4 and above retains 55\% at 93\%. Assigning the score in a dedicated second pass that re-reads each tag's quote, rather than inline during extraction, is what turns it into a usable dial: an inline self-rating is over-confident and close to a binary flag, placing three quarters of tags at the top score, while the second-stage re-grade spreads the same tags across the full range and reaches precision the inline score cannot. The economic evaluation is an event study over 182{,}174 cleanly attributable filing events using unsigned, variance-standardized abnormal returns; it makes no return-prediction claims. Within the single item code 5.02, reaction magnitudes differ systematically by tag: chief executive departures move prices by more than two standard deviations in 17\% of cases, against 10\% for routine officer appointments. Across the taxonomy, tag-level mean reactions span a threefold range, the most reactive tags are filed predominantly under the uninformative Item 8.01, and adding tags to item codes raises explained variance in reaction magnitude by 1.3 percentage points, roughly three times the 0.4-point increment from adding item codes to tags. The two evaluations connect through precision: the tags the judge flags as low-precision receive low quality scores, so filtering on the score removes them and leaves the reaction rankings essentially unchanged.

The contributions are fourfold. Firstly, a recipe for verifiable event extraction whose reliability does not rest on the model's own judgment: schema-constrained output and verbatim quote grounding make individual tags checkable against the filing, and a dedicated second pass that re-grades each cited quote assigns a calibrated quality score. The central methodological finding is that this second pass is what makes the score usable, since an inline self-rating saturates and behaves like a binary flag, while re-grading the same tags in isolation spreads them across the full range and reaches a precision the inline score cannot. Secondly, the calibrated score functions as a filtering dial: a downstream user trades precision against the share of tags retained by thresholding the score, with no change to the model or further runs over the corpus. Thirdly, evidence that SEC item codes are economically coarse and that fine-grained taxonomy tags carry relevant information item codes lack, is established by an event study that uses no language model. Fourthly, the full corpus of 601{,}088 grounded, quality-scored event tags over 292{,}984 filings is made available for further research.

%% file: sections/related.tex
\section{Related Work}
\label{sec:related}

Our work draws on three lines of research: the empirical literature that uses Form 8-K filings and their item codes, the application of language models to financial text, and the evaluation of language model outputs at scale.

\subsection{8-K filings and item codes}

The 2004 expansion of Form 8-K requirements established item codes as the standard taxonomy of material corporate events. Lerman and Livnat~\cite{lerman2010} document abnormal volume and return volatility around filings of both the newly required and pre-existing items, and subsequent work uses item codes and item counts as disclosure measures~\cite{he2020}. The item-code system has known limitations as an event-type label: voluntary items such as 2.02, 7.01, and 8.01 mix heterogeneous content~\cite{he2020}, and a single mandatory item can cover events with very different economic significance. Our contribution to this literature is direct measurement of that coarseness: we quantify the within-item heterogeneity of market reactions using labels finer than the item system, and we show that the most reactive event types reside disproportionately in the catch-all Item 8.01.

\subsection{Language models for financial text}

Textual analysis in finance progressed from dictionary methods~\cite{loughran2011,loughran2016} through pre-trained encoders fine-tuned for financial sentiment~\cite{araci2019} to large generative models trained or adapted for the financial domain~\cite{wu2023}. Corporate event detection from news text has been studied as a supervised task with fixed event vocabularies~\cite{zhou2021}, large filing corpora have been released for language model training~\cite{loukas2021}, and recent work documents that general-purpose language models extract economically meaningful signals from headlines~\cite{lopezlira2023}; augmented LLM pipelines with dedicated validation layers have likewise been used to extract structured, company-level signals such as tickers and sentiment from financial news~\cite{dolphin2024extracting}. Closest to our setting, taxonomy-aligned extraction has been applied to 10-K risk factors by pairing quote-grounded LLM extraction with embedding-based mapping to a taxonomy and LLM-as-a-judge filtering of spurious assignments~\cite{dolphin2026taxonomy}. Relative to this line, our system tags discrete corporate events under a three-tier taxonomy an order of magnitude finer than typical event-detection vocabularies, attaches structural validity guarantees (schema constraints~\cite{willard2023,geng2023} and verbatim quote grounding) to every extracted label rather than relying on the model's unconstrained output, and replaces post-hoc judge filtering with a calibrated quality score assigned by a dedicated grounding pass, which we validate economically through an event study.

\subsection{Evaluating model-generated labels}

Language models are increasingly used as annotators~\cite{gilardi2023}, although annotation studies emphasize that model labels require task-specific validation~\cite{pangakis2023}, and as judges of other models' outputs~\cite{zheng2023}, where self-preference toward their own generations is a documented bias~\cite{panickssery2024}. Two known failure surfaces motivate our evaluation design: generative models fabricate content unsupported by their inputs~\cite{ji2023}, and self-reported confidence is informative but requires empirical validation~\cite{lin2022,tian2023}. Where such confidence is typically elicited inline with the answer, we instead assign it in a dedicated pass that re-reads only the cited evidence against the category definition, which we find yields a sharper and better-calibrated dial. We combine an LLM-judge audit, stratified across event types and confidence levels, and equipped with verdicts that separate category errors from fabrications, with an economic validation in the event-study tradition~\cite{mackinlay1997,boehmer1991}. Using market reactions as an external check on label quality, and using label confidence to sharpen the event study, ties the two evaluation surfaces together in a way that neither provides alone.

%% file: sections/method.tex
\section{Extraction System}
\label{sec:method}

The extraction system assigns each filing a set of event tags drawn from a fixed taxonomy, with every tag carrying a supporting quote and a quality score. Tagging proceeds in two stages: an extraction stage that identifies events and cites a supporting quote for each, and a scoring stage that assigns each tag its quality score by judging the cited quote against the category definition. This section describes the taxonomy, the two stages, and the reliability mechanisms applied to the model's output.

\subsection{Taxonomy}
\label{sec:taxonomy}

The taxonomy is a three-tier hierarchy with 8 primary categories, 29 secondary categories, and 119 tertiary event types, written to describe economic events rather than filing obligations. The primary tier separates broad domains (leadership and governance, financial results, strategic transactions, capital and financing, operations and strategy, risk events, regulatory and compliance, and shareholder activity). The tertiary tier distinguishes, for example, the departure of a chief executive from that of another named officer, and a merger agreement from a merger completion or a deal withdrawal. Each tertiary type has a one- to two-sentence definition, and the model works from these definitions, not the type names alone. The taxonomy is organized by what happened, not by SEC item number, so a given event type can surface under any item.

\subsection{Two-stage extraction and reliability mechanisms}
\label{sec:agent}

\paragraph{Extraction (first stage).}
The first stage runs as a single model call per filing. The full taxonomy, with definitions, is placed in the prompt together with the filing text, and the model returns a list of tags, each consisting of a tertiary type and a supporting quote. We use a compact, commercially available instruction-tuned model at temperature zero; the design goal is that reliability comes from the surrounding structure rather than from model scale, and the evaluation in~\Cref{sec:intrinsic} tests that goal directly. Two checks run on the model's output at this stage.

\paragraph{Constrained output.}
Each tag is checked against the output schema: a tag whose event type is not an exact taxonomy entry is rejected, the error is returned to the model, and the model retries, up to three times. This guarantees that every tag the system emits is a valid taxonomy entry, without constraining the model's output token by token as it generates~\cite{willard2023}.

\paragraph{Quote grounding.}
Every tag must cite a supporting quote from the filing. The validator checks that quote against the filing text using four-word sequences: it breaks the quote into overlapping runs of four consecutive words and accepts the quote only when at least 40\% of those runs appear verbatim in the filing, triggering a retry otherwise. This makes a fabricated quote structurally impossible: a tag survives only if its cited text genuinely appears in the filing, aside from minor differences in spacing or punctuation. It does not, on its own, guarantee that the quoted text actually establishes the tagged event; \Cref{sec:intrinsic} measures that remaining gap between quoting the filing and supporting the event.

\paragraph{Scoring (second stage).}
The quality score is assigned in a separate stage. For each extracted tag, a separate grader model receives only the tag's cited quote and its category path with the category definition, not the rest of the filing. It rates, from 1 to 5, how well the quote on its own establishes that the tagged event occurred and that the category fits: a 5 requires the quote to establish the event explicitly and unambiguously with an exact category match, down to a 1 when the quote does not establish the event or points to a different category. Restricting the grader to the cited quote turns scoring into an explicit grounding check, the natural complement to the first stage's verbatim-quote requirement. Assigning the score this way, rather than having the extraction model rate its own output while generating it, is what makes the score a calibrated reliability dial; \Cref{sec:intrinsic} reports the calibration and an ablation against scoring inline during extraction. The quality score used throughout this paper is this second-stage score. It is attached to every tag, so downstream consumers can trade coverage against precision without rerunning either stage.

As a concrete example, a 2024 filing announcing the departure of a company's chief executive officer, effective the previous day, yields a tag with tertiary type \emph{CEO departure} and the supporting quote ``announced the departure of [the executive], the Company's Chief Executive Officer (`CEO') and Chairperson of the Company's Board of Directors''. The first stage verifies the quote against the filing text; the second stage, reading only this quote against the CEO-departure definition, assigns it a score of 5.

Filings whose parsed text exceeds roughly 150{,}000 characters, almost always large exhibit attachments rather than disclosure text, are excluded; and when the model finds no taxonomy-relevant event in a filing, it records an explicit no-match rather than forcing a tag.

%% file: sections/data.tex
\section{Data}
\label{sec:data}

We combine three data sources: tagged 8-K filings, SEC item codes extracted from the same filings, and daily market data.

\paragraph{Filings and tags.}
The corpus contains 292{,}984 Form 8-K filings by 9{,}669 distinct filers from January 2022 through June 2026, with parsed plain text of the item sections of each filing
\footnote{The cleaned 8-K text is available at \href{https://massive.com/docs/rest/stocks/filings/8-k-text?utm_source=research&utm_campaign=8k_tags}{massive.com/docs/rest/stocks/filings/8-k-text}}
. Running the extraction system over the corpus yields 608{,}346 output records, of which 7{,}258 are explicit no-match records covering 6{,}016 filings; the remaining 601{,}088 validated tags cover 286{,}968 filings (mean 2.1 tags per filing).
\footnote{The processed dataset and taxonomy are available at:
\\
\href{https://massive.com/docs/rest/stocks/filings/8-k-disclosures?utm_source=research&utm_campaign=8k_tags}{massive.com/docs/rest/stocks/filings/8-k-disclosures}.}
The quality scores are spread across the full range: 34\% of tags carry score 5, 21\% score 4, and 45\% scores 1 through 3.

\paragraph{Item codes.}
We read the SEC item codes directly from the filing text by matching the 32 legal item codes where they appear as line headers, recovering items for 99.8\% of filings. The resulting distribution matches the known item-code frequencies in EDGAR: Item 9.01 (exhibits) appears in most filings, followed by Items 2.02, 8.01, 7.01, and 5.02. Item 9.01 is excluded from all item-based analyses since it marks attached exhibits rather than an event.

\paragraph{Market data.}
We use split-adjusted daily open, high, low, close, and volume data for all U.S. listed and over-the-counter stocks from December 2021 through June 2026, and map each filer to its common-stock listing as of the filing month using reference data as it stood at the time, so the mapping uses no later information. Filings by entities without a common-stock listing (for example, debt-only issuers and trusts) are excluded from market analyses; secondary listings such as exchange-traded notes and preferred shares are remapped to the issuer's primary common stock. The mapped sample covers 243{,}069 filing events, of which 216{,}910 satisfy the baseline-window, price, and recent-trade filters described in~\Cref{sec:market}.

%% file: sections/intrinsic.tex
\section{Intrinsic Evaluation}
\label{sec:intrinsic}

The intrinsic evaluation asks two questions: are the tags correct, and is the quality score informative? We answer both with a language model judge applied to a stratified sample, and we check that the judge's verdicts are stable.

\subsection{Design}
\label{sec:judge-design}

We sample 5{,}125 tags stratified by both event type and quality score: up to nine tags from each of the 119 tertiary event types at each of the five scores, so that every score level is represented by roughly a thousand tags. Because the scores are sampled in equal numbers, the sample over-represents low-scoring tags relative to how often they occur in the full tag set. Every aggregate precision we report is therefore reweighted to the full tag set's score distribution: we measure precision within each score and recombine the scores in the proportions they occur across all tags. The per-score numbers reported below are conditional on the score and need no such reweighting.

For each tag the judge receives the full filing text, the tag's three-level category path with its definition, and the supporting quote cited for the tag. It returns one of three verdicts: \emph{correct} (the filing contains the event and the category is a reasonable reading of it), \emph{clearly wrong} (the event does not belong to the category under any reasonable reading), and \emph{not supported} (the filing does not contain the claimed event). The judge is Opus 4.8, a more capable model than the extraction model. Because the judge sees only the definition of the single category under evaluation, not the full 119-entry taxonomy, it judges whether that one category applies rather than searching for a better label. The bar is still strict: a tag is marked wrong when its category does not apply even though a closely related event is present, for instance a guidance-withdrawal tag on a filing that revises rather than withdraws guidance, or a spin-off-completion tag on a spin-off that has only been signed. The resulting precision figures are correspondingly conservative.

Separating the unsupported verdict matters. A binary correct/incorrect judge conflates two very different failures: a tag whose claimed event is real but does not fit the assigned category, and a tag for an event that never happened. We therefore report \emph{precision} (the share of tags judged correct) and the \emph{unsupported rate} (the share whose claimed event is absent from the filing) separately throughout; precision captures whether the tagged event is real and reasonably categorized, while the unsupported rate isolates outright hallucination, where the filing does not contain the claimed event at all.

\subsection{Precision and calibration}
\label{sec:calibration}

Precision over all tags is 0.726, with 1.6\% of tags not supported by the filing text. Conditioning on the quality score sharpens this. Keeping only the top score retains 34\% of tags at 0.964 precision with essentially no unsupported tags; score 4 and above retains 55\% at 0.927 precision and 0.1\% not supported; score 3 and above retains 61\% at 0.912. In practice, the quality score exists as a filtering mechanism, and any downstream use case of the tags would restrict to those above a chosen threshold. As a result, the score 4 and score 5 precision numbers are the most indicative. The category errors the judge does flag are typically near-misses rather than unrelated mislabels: a real event is present but the assigned category does not quite fit, as when an accounting-error-correction tag is applied to what is actually a full financial restatement, or a class-action tag to individual rather than class-action shareholder suits.

\begin{table}[t]
\caption{Judge-assessed precision by quality score. The unsupported rate is the share of tags whose claimed event is absent from the filing. Counts are over the evaluation sample, which is balanced across quality scores and so over-represents low-scoring tags relative to the full tag set.}
\label{tab:calibration}
\begin{tabular}{rrrr}
\toprule
Score & $n$ & Precision & Unsupported \\
\midrule
1 & 1{,}013 & 0.123 & 0.080 \\
2 & 1{,}051 & 0.472 & 0.033 \\
3 & 981 & 0.782 & 0.015 \\
4 & 1{,}041 & 0.868 & 0.002 \\
5 & 1{,}039 & 0.964 & 0.000 \\
\bottomrule
\end{tabular}
\end{table}

\begin{figure}[t]
\centering
\includegraphics[width=\columnwidth]{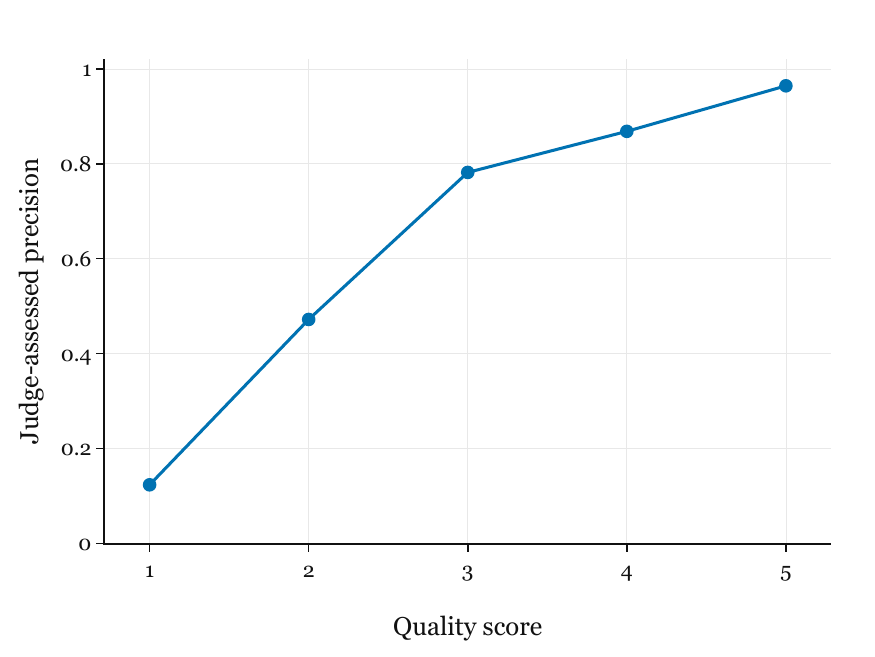}
\caption{Calibration of the quality score: judge-assessed precision by score, where precision is the share of tags the judge finds correct (the event is real and the tag is a defensible reading of it).}
\label{fig:calibration}
\end{figure}

\Cref{tab:calibration} and~\Cref{fig:calibration} show the central result of this section: judge-assessed precision rises monotonically with the quality score, from 0.123 at score 1 to 0.964 at score 5. We use the word \emph{calibration} loosely here, since the score is not a probability and standard probabilistic calibration metrics~\cite{guo2017} do not apply; we mean only that precision increases with the score. The score therefore works as a tunable reliability dial: keeping only the top score yields 96\%-precision labels on 34\% of the corpus, and score 4 and above yields 93\% on 55\%, with no separate filtering model. This extends work on verbalized confidence~\cite{lin2022,tian2023} to a constrained extraction setting, where the score is produced by a dedicated quote-grounding pass and attached to each structured label rather than to a free-form answer.

Hallucination, a tag whose claimed event is absent from the filing altogether rather than a mis-categorized real event, is the rarer and more serious failure, and the quote grounding targets it directly: the first stage requires every tag to cite verbatim filing text, and the second stage re-reads that quote against the category definition. It is concentrated at the low scores and falls to near zero after filtering, from 8\% at score 1 to 0.2\% at score 4 and none of the score-5 tags in the sample. The small differences among the top scores amount to a handful of tags, and we do not read them as meaningful.

\begin{figure*}[t]
\centering
\includegraphics[width=\textwidth]{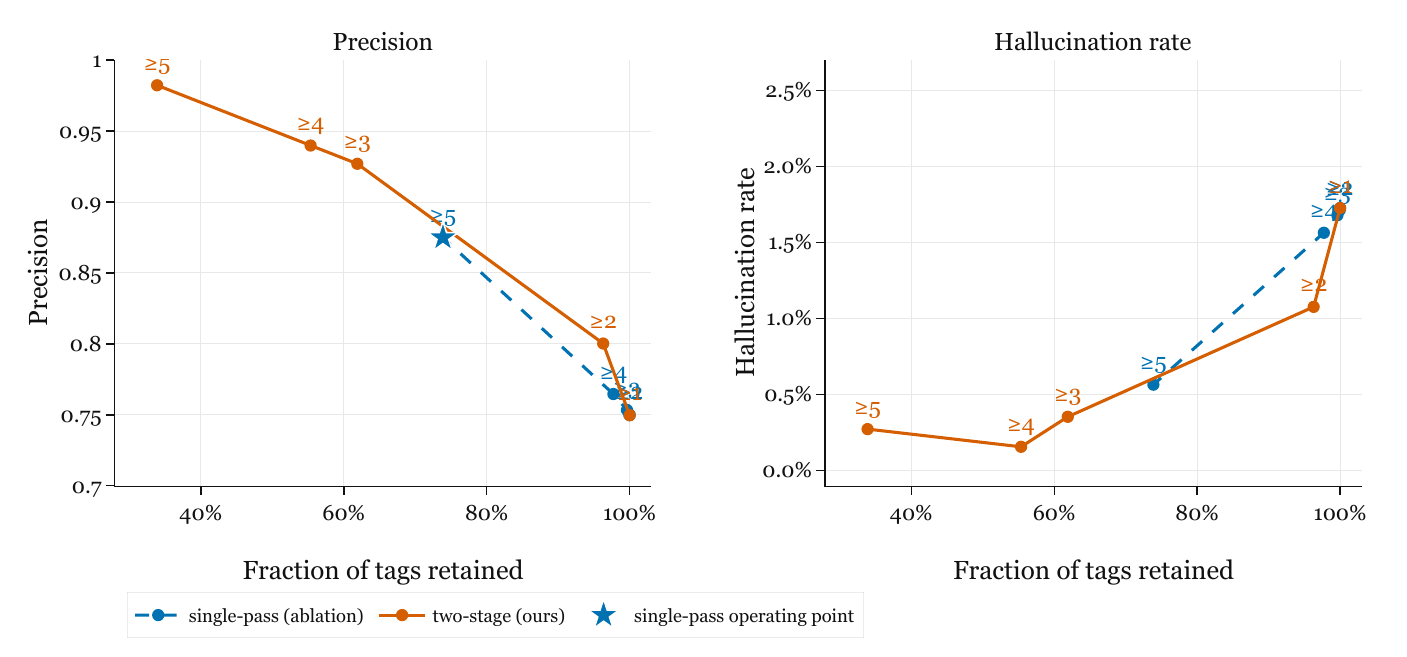}
\caption{Precision against the fraction of tags retained as the quality-score keep-threshold sweeps from 1 to 5. The second-stage score (solid) is a genuine dial: it reaches 0.94 precision at 55\% of tags and 0.98 at 34\%. Scoring inline during a single extraction pass (dashed) saturates: its most selective point (star) still retains 73\% of tags, at 0.875 precision, and cannot reach higher precision at any threshold. The right panel shows the hallucination rate over the same sweep: the two-stage score drives it below 0.4\% at its high-confidence thresholds, whereas the inline score floors near 0.6\% at its most selective point.}
\label{fig:regrade}
\end{figure*}

\paragraph{Why scoring is a separate stage.}
The calibration above depends on assigning the score in a dedicated pass. The alternative, having the extraction model rate each tag inline as it generates it, is markedly over-confident: it places roughly three quarters of all tags at the top score and almost all of the rest at score~4, leaving the lower three scores nearly empty. The inline score is therefore close to a binary flag (top score or not) rather than a graded dial. \Cref{fig:regrade} plots both dials as precision against the fraction of tags retained, on the original sample stratified on the inline score so that the inline dial is unbiased. Every keep-threshold below the maximum retains almost the entire corpus, so the inline operating points for scores~1 through~4 collapse together and only the top-score cut separates: the inline score's one informative operating point still retains 73\% of all tags, at 0.875 precision, with no way to trade coverage for higher precision beyond that.

The second-stage re-grade instead redistributes the same tags across all five score levels, with operating points spanning 34\% to 100\% of the corpus: it is a genuine dial whose precision climbs to 0.94 at 55\% of tags and 0.98 at 34\%, an operating range the inline score cannot reach at any threshold. The mechanism is visible inside the inline top score: re-grading those tags from their quotes alone isolates a pocket, scored 1 by the second stage, that is only 14\% precise with 14\% unsupported, exactly the over-confident tags that inline scoring cannot flag.

Three caveats bound the score. Firstly, the second-stage grader is drawn from the same model family as the extraction model, so the score is a form of self-assessment; its predictive value is nonetheless measured against the independent and more capable judge, not against the grader agreeing with itself. Secondly, the grader sees only the cited quote, which keeps the check cheap and tests quote-grounding directly but leaves it blind to the rare case in which a quote reads correctly in isolation while the surrounding filing contradicts it. Thirdly, higher precision comes at the cost of coverage, so the score is a tunable dial rather than a single operating point.

%% file: sections/market.tex
\section{Economic Validation}
\label{sec:market}

The economic evaluation asks whether the taxonomy separates events the market treats differently, and whether it does so beyond what item codes already capture. We measure only the size of the price move around each filing, not its direction, and we make no attempt to predict future returns.

\subsection{Design}
\label{sec:event-design}

For each filing we measure how much the filer's stock moved around the filing, in either direction, following standard event-study practice~\cite{boehmer1991,mackinlay1997}. Let $r_{i,t}$ be the stock's close-to-close return and $r_{m,t}$ the return of a broad market exchange-traded fund; the market-adjusted return is $a_{i,t} = r_{i,t} - r_{m,t}$. To make stocks comparable, we take its absolute value and divide by the stock's normal daily variation, giving the reaction measure $\mathrm{SAR}_{i,t} = |a_{i,t}| / \hat\sigma_i$, where $\hat\sigma_i$ is the standard deviation of $a_{i,t}$ over a baseline window of 60 trading days ending 11 trading days before the filing; the eleven-day gap keeps any run-up just before the filing out of the baseline. A value of 2, for instance, means the stock moved twice as much as it usually does in a day.

Filing records carry only the date, not the time, so a filing submitted after the market closes moves prices the next day. We therefore set the event day $t_0$ to the first trading day on or after the filing date, and use the larger of the two reactions on $t_0$ and the following trading day. A filing with no real news would average about 1.0 on this two-day measure (a single day would average about 0.8); we draw a dashed line at 1.0 in every figure as a no-news benchmark, so points above it mark larger-than-normal moves.

We keep an event only if the stock has at least 40 trading days of history in the baseline window, traded above one dollar before the filing, and last traded within five calendar days of the event day (otherwise its apparent one-day return spans a longer no-trade gap). These conditions leave 216{,}910 events. We add one further requirement, that no other filing by the same issuer fall within one trading day, so that a single price move is never credited to two filings; this leaves 182{,}174 events.

To attribute a price move to a specific event type, the filing must not be about several things at once. For the tag-level analyses we therefore keep only filings with a single substantive item code (ignoring the exhibits item), 125{,}726 events, and group them by which tags they carry. We do not instead require a single tag, because that would skew the sample toward routine filings: major events usually come with related tags (a chief executive departure is typically filed alongside the successor's appointment).

\subsection{Heterogeneity within one item code}
\label{sec:within-item}

\begin{figure}[t]
\centering
\includegraphics[width=\columnwidth]{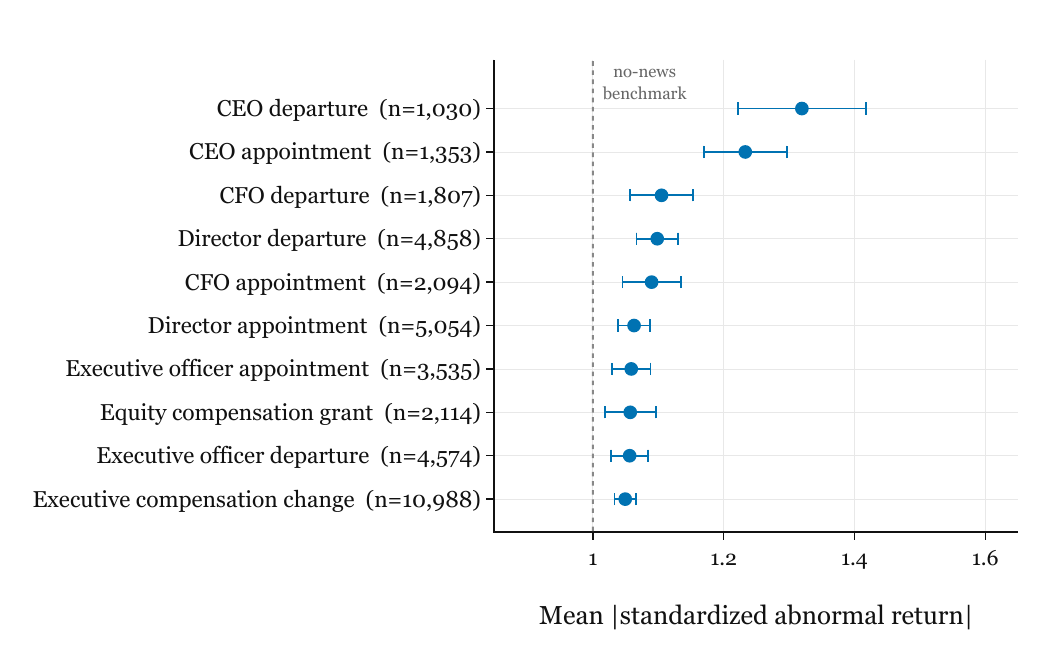}
\caption{Mean reaction size ($\mathrm{SAR}$) for filings whose only substantive item is 5.02, grouped by taxonomy tag, with 95\% confidence intervals. The dashed line marks the no-news benchmark.}
\label{fig:502}
\end{figure}

Item 5.02 (departures and appointments of directors and officers) is the clearest example of how coarse item codes are. \Cref{fig:502} splits the 19{,}443 filings whose only substantive item is 5.02 by taxonomy tag, and the reactions differ substantially from one tag to the next (Kruskal--Wallis $p = 3 \times 10^{-8}$). Chief executive departures produce an average reaction of 1.32 (the stock moves about 1.3 times its normal daily amount, versus 1.0 for a filing with no news), with 17.2\% of these filings moving more than twice their normal amount and 3.5\% more than four times. Routine officer appointments produce only 1.06, with 9.7\% and 0.8\%. That roughly fourfold gap at the four-times mark separates the two tags more clearly than the averages alone, yet an item-code user sees every one of these filings under the single label 5.02. This Kruskal--Wallis test and the one in~\Cref{sec:robustness} are the only formal significance tests in the paper; both stay significant even after accounting for the many tags being compared at once, and the error bars in the figures are shown for visual context, not as significance tests.

\subsection{Heterogeneity across the taxonomy}
\label{sec:across-tags}

\begin{figure}[t]
\centering
\includegraphics[width=\columnwidth]{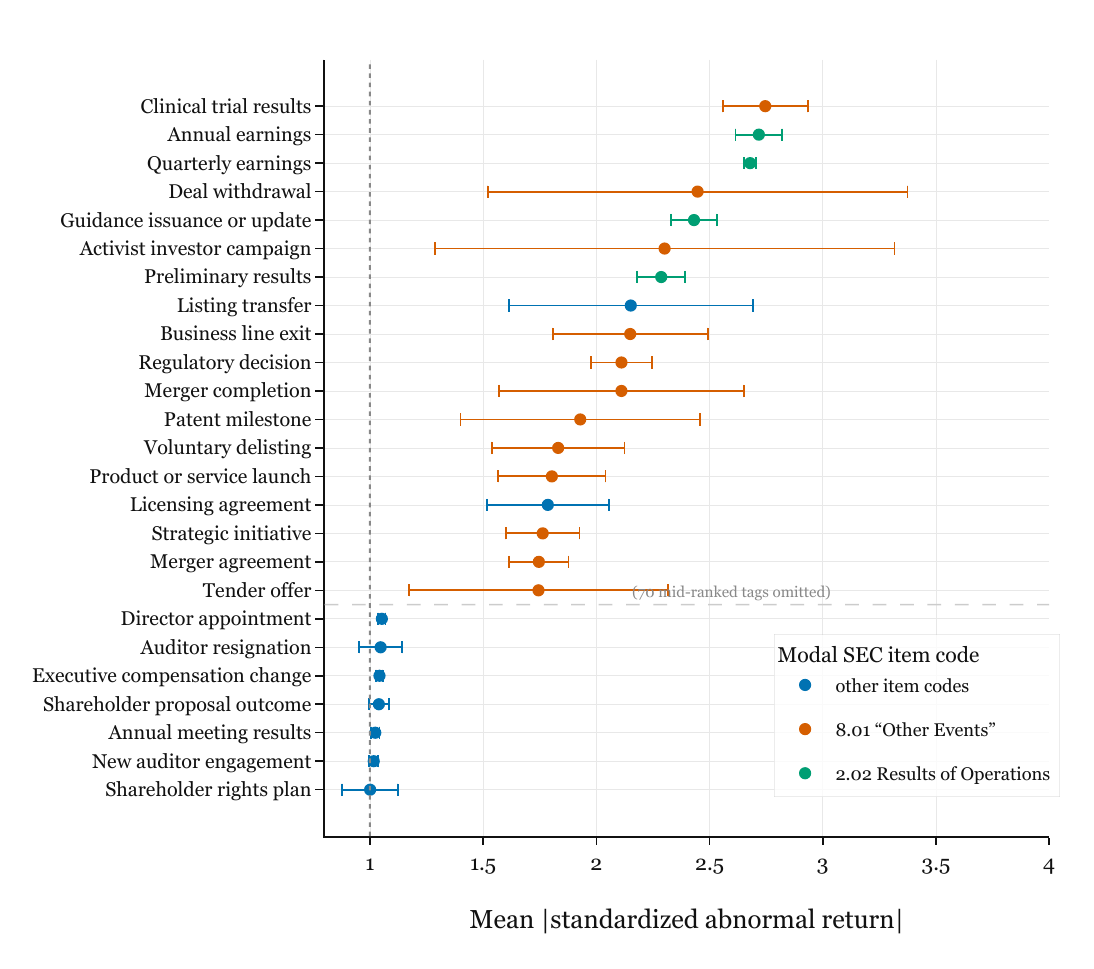}
\caption{Mean reaction size ($\mathrm{SAR}$) by taxonomy tag (top and bottom of the ranking; tags with at least 150 events), with 95\% confidence intervals, colored by the tag's most common item code. The dashed line marks the no-news benchmark.}
\label{fig:tagrange}
\end{figure}

Across all tags with at least 150 cleanly attributed events, the most reactive tags exhibit roughly three times the average reaction of the least reactive (\Cref{fig:tagrange}): from clinical trial results at 2.75 (with 34\% of these filings moving more than twice their normal daily amount) and earnings releases around 2.7, down to auditor engagements and annual-meeting results close to the 1.0 no-news benchmark. Two features of this ranking bear directly on the item-code question. Firstly, routine event types fall almost exactly on the no-news line, a useful check on the measure. Secondly, nine of the fifteen most reactive tags have Item 8.01, the voluntary ``Other Events'' catch-all, as their most common item code: clinical trial results, merger completions and agreements, deal withdrawals, regulatory decisions, voluntary delistings, and patent milestones are filed mostly under the one item that carries no event-type information at all.

We can also measure how much of the variation in reaction size each labeling scheme accounts for. Item codes alone explain 15.0\%; tags alone explain 15.9\%; the two together explain 16.3\%. Adding tags on top of item codes raises the explained variance by 1.30 percentage points, while adding item codes on top of tags raises it by only 0.41, a ratio of about three (95\% bootstrap confidence interval 2.6 to 4.0, with the tag increment exceeding the item increment in every resample). The tags thus capture nearly all of the reaction-relevant information in item codes, and the converse does not hold. Both increments are small relative to the total, as expected for the size of cross-sectional price moves; the comparison of interest is between the two. The analysis so far used only the 125{,}726 single-item filings, which are simpler than a typical 8-K. To check that the result is not an artifact of that restriction, we repeat it on all 182{,}174 collision-free filings, now including the multi-item ones, encoding each filing's full set of items and tags. The tag increment remains the larger, by a wider margin (1.86 points against 0.55), so the finding does not depend on the single-item restriction.

\subsection{Confidence filtering as robustness}
\label{sec:robustness}

The intrinsic evaluation flagged some tags as imprecise, which raises the concern that mistagged filings might be driving these results. Repeating the analyses on only the tags scored 4 and above, the roughly 93\%-precision subset, leaves every conclusion intact: the tag ranking is essentially unchanged (Spearman correlation 0.939, with 10 of the top 15 unchanged), the Item 5.02 gradient still holds (Kruskal--Wallis $p = 1 \times 10^{-7}$), eight of the top fifteen movers are still filed mostly under Item 8.01, and the variance asymmetry persists (the tag increment is 2.6 points against 1.9 for the converse, on the filings that retain a high-confidence tag).

The filter removes the tags the judge flagged as imprecise. The shareholder director-nomination tag, among the least precise in the taxonomy because it fires on procedural deadline notices, scores below 4 on 83\% of its instances and so largely drops out of the high-confidence subset; the second-stage grader recognizes that a procedural notice does not establish a real event. Genuine events are retained, with their reactions little changed: deal withdrawals, for example, keep an average reaction near 2.4. Filtering therefore stabilizes the economic results by removing imprecise tags rather than by altering the reactions of correctly tagged ones. This is the connection between the two evaluations: the same score that predicts label precision also marks the filings the event study should set aside.

Two features of the design make these estimates conservative. Stocks that do not trade on the event day, including halted stocks, drop out, which truncates the most extreme reactions. Because excluding same-day filing collisions lowered the measured reactions of routine tags, a simpler design that ignored collisions would overstate the differences between event types.

%% file: sections/exhibits.tex
\section{The Taxonomy as a Measurement Instrument}
\label{sec:exhibits}

Beyond the event study, the tagged corpus supports descriptive measurement that item codes cannot. This section presents three analyses chosen to probe different validity properties: tracking of a regulatory regime change, tracking of macroeconomic conditions, and cross-sectional coherence.

\subsection{Regulatory regime change: cybersecurity disclosure}
\label{sec:cyber}

\begin{figure}[t]
\centering
\includegraphics[width=\columnwidth]{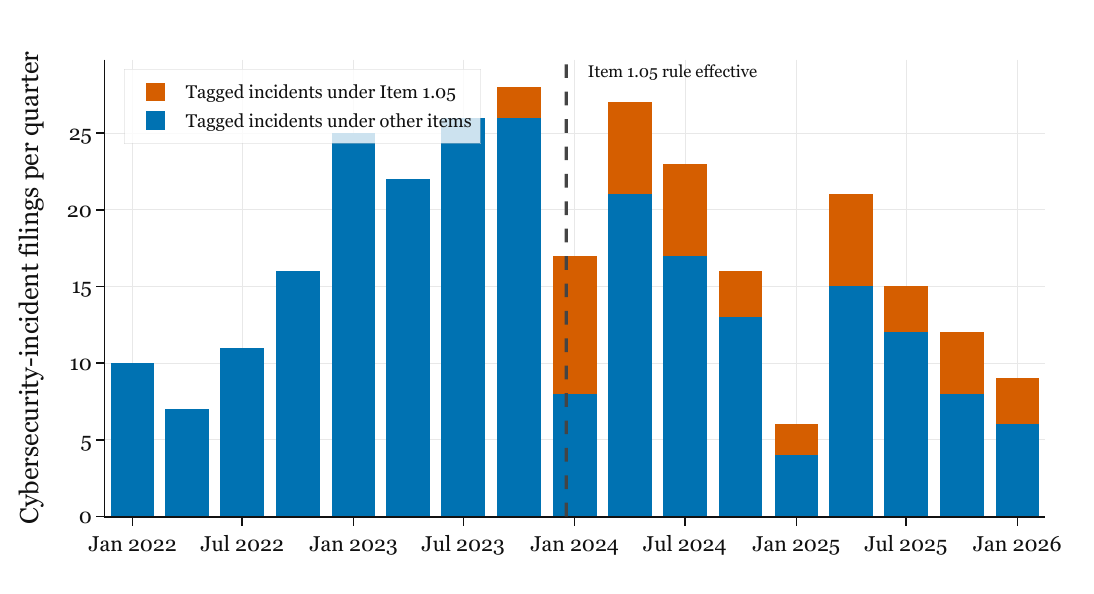}
\caption{Quarterly counts of filings tagged as cybersecurity incidents, split by whether the filing carries Item 1.05. The dashed line marks the effective date of the SEC incident-disclosure rule.}
\label{fig:cyber}
\end{figure}

In December 2023, a new SEC rule introduced Item 1.05, requiring disclosure of material cybersecurity incidents under a dedicated item code~\cite{sec2023cyber}. The taxonomy contains a cybersecurity-incident event type that is independent of item codes, which makes the rule a natural experiment in how well item codes measure an event class. The corpus contains 309 filings tagged as cybersecurity incidents, of which only 48 carry Item 1.05 (\Cref{fig:cyber}). Before the effective date the count under Item 1.05 is necessarily zero; after it, roughly two thirds of tagged incidents continue to be disclosed under other items, predominantly the catch-all Item 8.01. An item-code-based measure of cybersecurity incidents would have recorded nothing before 2024 and a minority of incidents after, while the content-based tag measures the event class continuously across the regime change.

\subsection{Macroeconomic narratives}
\label{sec:macro}

\begin{figure}[t]
\centering
\includegraphics[width=\columnwidth]{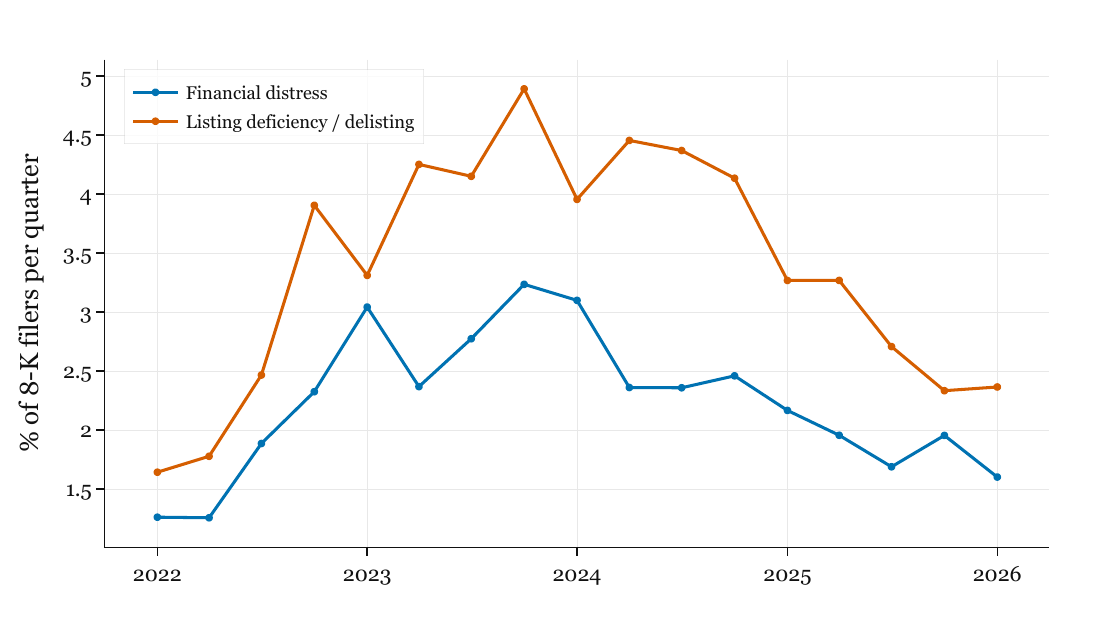}
\caption{Share of 8-K filers per quarter carrying financial-distress tags (going concern, covenant violation, payment default, bankruptcy, restructuring, impairment) and exchange-listing-pressure tags (deficiency notices, delisting determinations, voluntary delistings).}
\label{fig:macro}
\end{figure}

Aggregated tag frequencies track macroeconomic conditions with the timing one would expect (\Cref{fig:macro}). The share of filers carrying financial-distress tags rises from 1.3\% of filers in early 2022 to above 3\% through 2023 and early 2024, the period following the sharp rise in interest rates, and declines thereafter. Exchange-listing-pressure tags rise from 1.6\% to a peak of 4.9\% in late 2023 and halve by 2025, consistent with the post-2021 cohort of small listings exiting compliance. These series are byproducts of the event labels rather than purpose-built indicators, and their consistency with the known macroeconomic record is a corpus-level validity check.

\subsection{Cross-sectional coherence}
\label{sec:sector}

\begin{figure}[t]
\centering
\includegraphics[width=\columnwidth]{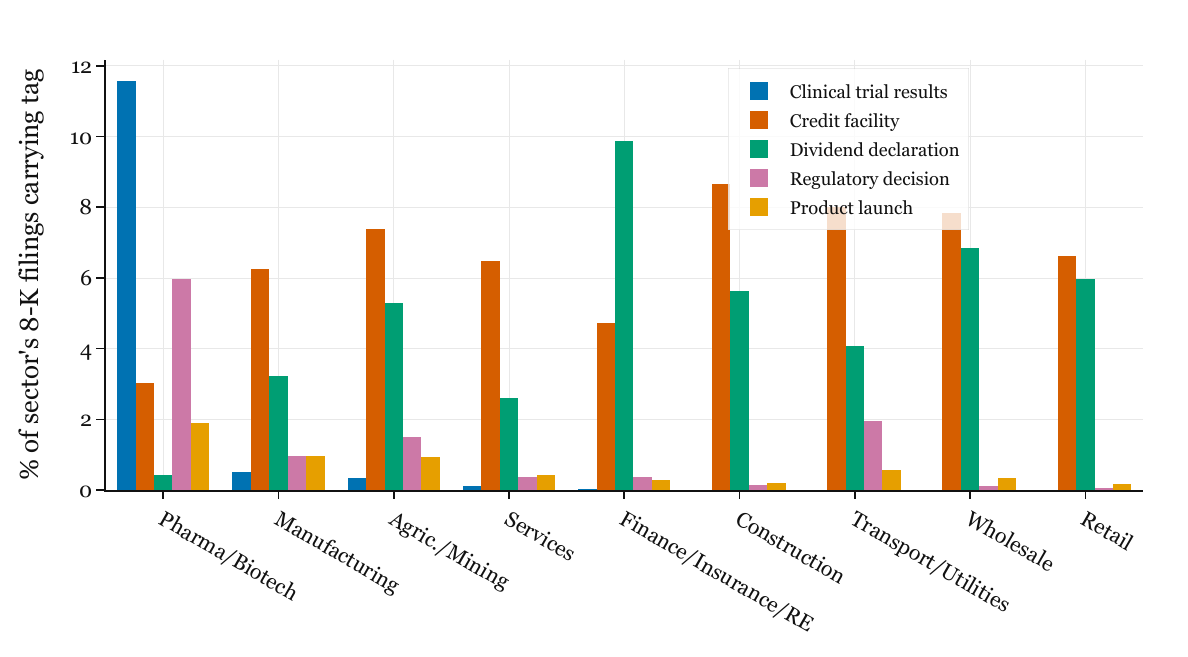}
\caption{Share of each sector's 8-K filings carrying selected tags. Sectors are SIC divisions, with pharmaceutical and biotechnology filers shown separately.}
\label{fig:sector}
\end{figure}

Tag incidence by industry behaves as domain knowledge requires (\Cref{fig:sector}). Clinical-trial-result tags appear on 11.6\% of pharmaceutical and biotechnology filings and essentially nowhere else; regulatory-decision tags concentrate in the same sector; dividend declarations concentrate among financial firms; credit-facility events are spread broadly across capital-intensive sectors. None of this information is recoverable from item codes, under which a clinical trial readout and a dividend declaration are both, typically, Item 8.01 filings.

%% file: sections/conclusion.tex
\section{Conclusion}
\label{sec:conclusion}

We have presented an LLM-based system that tags Form 8-K filings against a three-tier taxonomy of 119 event types using reliability mechanisms that do not rest on the model's own judgment. A first stage extracts tags under schema-constrained output and verbatim quote grounding, both checkable against the filing, and a second stage re-grades each tag's cited quote to assign a quality score from 1 to 5. Applied to 292{,}984 filings, the system produces 601{,}088 grounded tags, which we release. The intrinsic evaluation shows that the quality score is a usable filtering dial: judge-assessed precision rises monotonically from 12\% at the lowest score to 96\% at the top, so thresholding on the score trades the share of tags retained for a higher-precision label set, and the central methodological lesson is that assigning the score in a dedicated pass rather than inline during extraction is what makes the dial sharp. The economic evaluation, an event study that uses no language model, provides an external check on label quality and shows that the taxonomy separates events the market treats differently, carrying reaction-relevant information that SEC item codes lack.

These results come with limitations. Quote grounding prevents fabricated evidence but not fabricated inference, and the second-stage grader sees only the cited quote, so it cannot catch a quote that reads correctly in isolation while the surrounding filing contradicts it; this residual is small at high confidence (0.1\% unsupported at score 4 and above, with none observed at the top score in our sample) but cannot be assumed to be zero. The event study uses daily prices and filing dates without time of day, so reaction windows span two trading days and filings by halted or delisted stocks drop out, truncating the extreme tail of reactions; and the corpus begins in 2022, so the macroeconomic series cover a single rate cycle.

The methodology, an inexpensive structurally validated extractor audited by a stronger judge, with the audit findings checked against market data, does not depend on properties specific to the 8-K. We expect it to transfer to other classes of financial disclosure, and document processing more generally, which we leave to future work.